\documentclass[10pt,twocolumn,letterpaper]{article}

\usepackage{cvpr}              
\usepackage{algorithmic}
\usepackage[ruled]{algorithm2e}
\usepackage{graphicx}
\usepackage{subcaption}
\usepackage{afterpage}
\usepackage{wrapfig}
\usepackage[justification=centering]{caption}
\usepackage{graphicx} %
\usepackage{subcaption} %
\usepackage{float}  %
\usepackage[justification=raggedright]{caption} 
\DeclareUnicodeCharacter{2081}{\textsubscript{1}}

\definecolor{cvprblue}{rgb}{0.21,0.49,0.74}
\usepackage[pagebackref,breaklinks,colorlinks,allcolors=cvprblue]{hyperref}

\def\paperID{9054} 
\def\confName{CVPR}
\def\confYear{2025}

\title{Hyperspectral Image Cross-Domain Object Detection Method \\ based on Spectral-Spatial Feature Alignment}

\author{
Hongqi Zhang$^{1}$ \quad
He Sun$^{2}$ \quad
Hongmin Gao$^{1}$ \quad
Feng Han$^{2}$ \quad
Xu Sun$^{2}$ \quad
Lianru Gao$^{2}$\thanks{Corresponding author: gaolr@aircas.ac.cn} \quad
Bing Zhang$^{2}$ \\
$^1$Hohai University, Nanjing, China \\
$^2$Aerospace Information Research Institute, Chinese Academy of Sciences, Beijing, China \\
{\tt\small zhqdmu@gmail.com, sunhe@aircas.ac.cn, gaohongmin@hhu.edu.cn, hfwindy@foxmail.com} \\
{\tt\small sunxu@aircas.ac.cn, gaolr@aircas.ac.cn, zb@radi.ac.cn}
}

\begin{document}
\maketitle
\begin{abstract}
With consecutive bands in a wide range of wavelengths, hyperspectral images (HSI) have provided a unique tool for object detection task. However, existing HSI object detection methods have not been fully utilized in real applications, which is mainly resulted by the difference of spatial and spectral resolution between the unlabeled target domain and a labeled source domain, i.e. the domain shift of HSI. In this work, we aim to explore the unsupervised cross-domain object detection of HSI. Our key observation is that the local spatial-spectral characteristics remain invariant across different domains. For solving the problem of domain-shift, we propose a HSI cross-domain object detection method based on spectral-spatial feature alignment, which is the first attempt in the object detection community to the best of our knowledge. Firstly, we develop a spectral-spatial alignment module to extract domain-invariant local spatial-spectral features. Secondly, the spectral autocorrelation module has been designed to solve the domain shift in the spectral domain specifically, which can effectively align HSIs with different spectral resolutions. Besides, we have collected and annotated an HSI dataset for the cross-domain object detection. Our experimental results have proved the effectiveness of HSI cross-domain object detection, which has firstly demonstrated a significant and promising step towards HSI cross-domain object detection in the object detection community.
\end{abstract}    
\section{Introduction}
\label{sec:intro}

Hyperspectral Images (HSI) \cite{cvpr_hsi, BHIS, Wang_2024_CVPR, Li_2022_CVPR, Li_2023_CVPR} can distinguish different materials with the aid of spectral information contained in consecutive bands \cite{Zhang_2023_CVPR, Spp_science}, and it has shown increasing application values in the hyperspectral object detection task (HOD). However, the existing HOD methods \cite{YunsongLi} have not yet been effectively utilized in practical applications, the main reason is that the domain shift problem between the labeled source domain and unlabeled real scenario, i.e., the target domain. Accordingly, this paper aims to explore the HSI cross-domain object detection (HCOD) for the first time.


\begin{figure}[t]
\centering
\begin{subfigure}{0.45\linewidth}
    \centering
    \includegraphics[width=\linewidth]{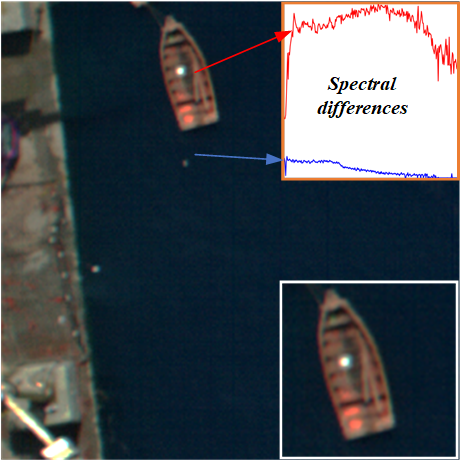}
    \caption{The difference between object and its surrounding background in the source domain}
\end{subfigure}
\hspace{0.05\linewidth} %
\begin{subfigure}{0.45\linewidth}
    \centering
    \includegraphics[width=\linewidth]{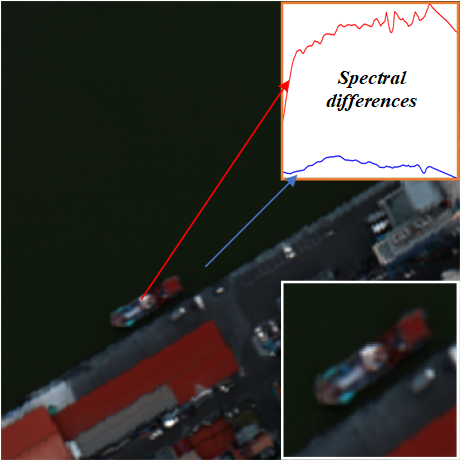}
    \caption{The difference between object and its surrounding background in the target domain}
\end{subfigure}

\vspace{0.5em} %

\begin{subfigure}{0.45\linewidth}
    \centering
    \includegraphics[width=\linewidth]{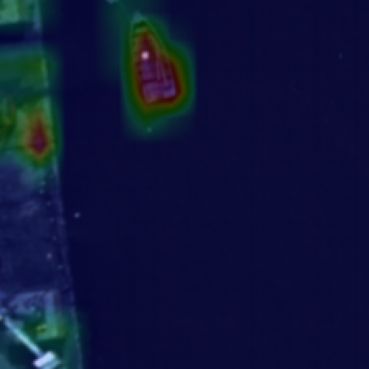}
    \caption{The extracted local spectral-spatial feature in the source domain}
\end{subfigure}
\hspace{0.05\linewidth} %
\begin{subfigure}{0.45\linewidth}
    \centering
    \includegraphics[width=\linewidth]{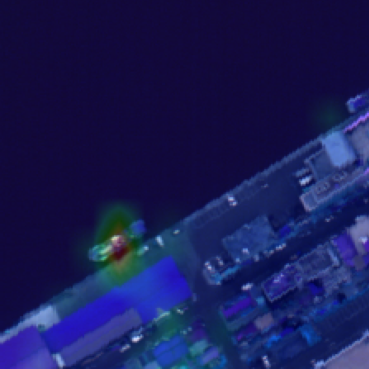}
    \caption{The extracted local spectral-spatial feature in the target domain}
\end{subfigure}

\caption{Illustration of the invariant local spatial-spectral characteristics and extracted corresponding features from SFA.}
\label{fig_HCOD}
\end{figure}

The unsupervised cross-domain object detection (UCOD) \cite{MeGA-CDA, SIGMA, TDD, SC-UDA, ConfMix, OnlineDA} has received great attention \cite{MGADA, AT, Dafaster, HT, MRT, PT} in the past years as there is usually no prior information about the target domain. To mitigate the domain shift problem, some work \cite{DThreeT, Dafaster, MGADA, MRT, AT, TIA, DDF, D_adapt, IIDA, YOLODA, Meta-UDA, ADA} have adopted the gradient reversal layer (GRL) \cite{GRL} to multiply represents for different domains, such as global-level \cite{Dafaster}, local-level \cite{HT}, instance-level \cite{Dafaster}, and pixel-level \cite{MGADA} representation. In addition, many researchers have designed various frameworks based on the teacher-student model \cite{PT, MRT,AT, HT,UMT,CMT, 2PCNet}, which has been widely applied to the UCOD task.

Although current UCOD have achieved success \cite{CRDA, GPA, HTCN, CDN, EPMDA, PDAHR, PDA} on most RGB images, they fail to apply in the HCOD task. The domain shift problem of HCOD is presented in both the spatial and spectral dimension. Therefore, it is necessary to align the spectral-spatial features from both domains in the HCOD task. Accordingly, the invariant spectral-spatial features in both domains are desired for achieving a promising performance of HCOD. As shown in Figure \ref{fig_HCOD}, the difference of spectral curves between the object and its local background in both domains are rather similar, and objects with a same semantic class usually share a similar local spatial information in both domain. Hence, the local spectral-spatial features are invariant between domains, it is necessary to extract and align this kind of features of both domains to achieve the HCOD.

According to the above analysis, it is necessary to capture the invariant spectral-spatial characteristics for HCOD by extracting the local spectral-spatial features. As our aim is to achieve HCOD in an unsupervised manner, it is more suitable to extract these features by deep network using a self-supervised training signal. Generally, an autoencoder (AE)-based module \cite{autoconder} can automatically transform the high-dimensional HSI data into a low-dimensional hidden space with the self-supervised reconstruction signal, where the learned hidden features reserve most of the local spectral-spatial features of HSI. Therefore, the AE-based reconstruction module can be viewed as an simple yet effective tool to capture the local spectral-spatial features.

Motivated by that, we have proposed a HCOD method based on spectral-spatial feature alignment (SFA). Our method has designed a more suitable AE-based architecture as the backbone of HCOD, namely the spectral-spatial alignment module (SSAM), which can effectively capture the local spectral-spatial features. On this basis, we have integrated a GRL in the domain classifier, which aim to align local spectral-spatial features between domains. In addition, a spectral autocorrelation module (SACM) has been developed on the aligned local spectral-spatial features specifically, which performs a key role in solving the domain shift in spectral dimension. Besides, we have collected and annotated an HSI dataset specifically for HCOD. The contributions of this manuscript can be summarized as follows:
\begin{enumerate}[]
\item To the best of our knowledge, we have first introduced an HCOD framework by aligning spectral-spatial features between the source and target domain, namely SFA.
\item We propose spectral-spatial alignment and spectral autocorrelation modules to solve the problem of domain shift in both spectral and spatial dimensions.
\item We have collected and annotated a dataset for the HCOD task. Our experimental results have shown that the proposed SFA has a superior performance over the state-of-the-art UCOD methods.
\end{enumerate}
\section{Previous Work}
\label{sec:formatting}
\subsection{HSI Object Detection}
HOD aims to locate and identify the object with the aid of spectral discriminative information. The existing methods can be divided into traditional-based and deep learning-based \cite{HTD1,HTD2, HTD3}. Most traditional-based methods obtain the HOD results by comparing the similarities between the spectral curves of detected pixel and the prior spectral curve (PSC), the most represented methods are constrained energy minimization \cite{CEM} and the adaptive cosine estimator \cite{ACE}. However, they heavily rely on the quality of PSC and suffer from the computational burden of processing large-scale datasets. In contrast, the deep learning-based method can detect the desired objects with a deep network supervised by a large set of training samples over the PSC, which can significantly improve the effectiveness and robustness of HOD. However, deep learning-based methods are rare due to the lack of HOD datasets. Jang et al. have proposed Double FPN-based network \cite{NIPS} for HOD, which achieved a promising performance with the aid of deep features.

\begin{figure*}[htbp]
\centering
\includegraphics[width=0.7\linewidth]{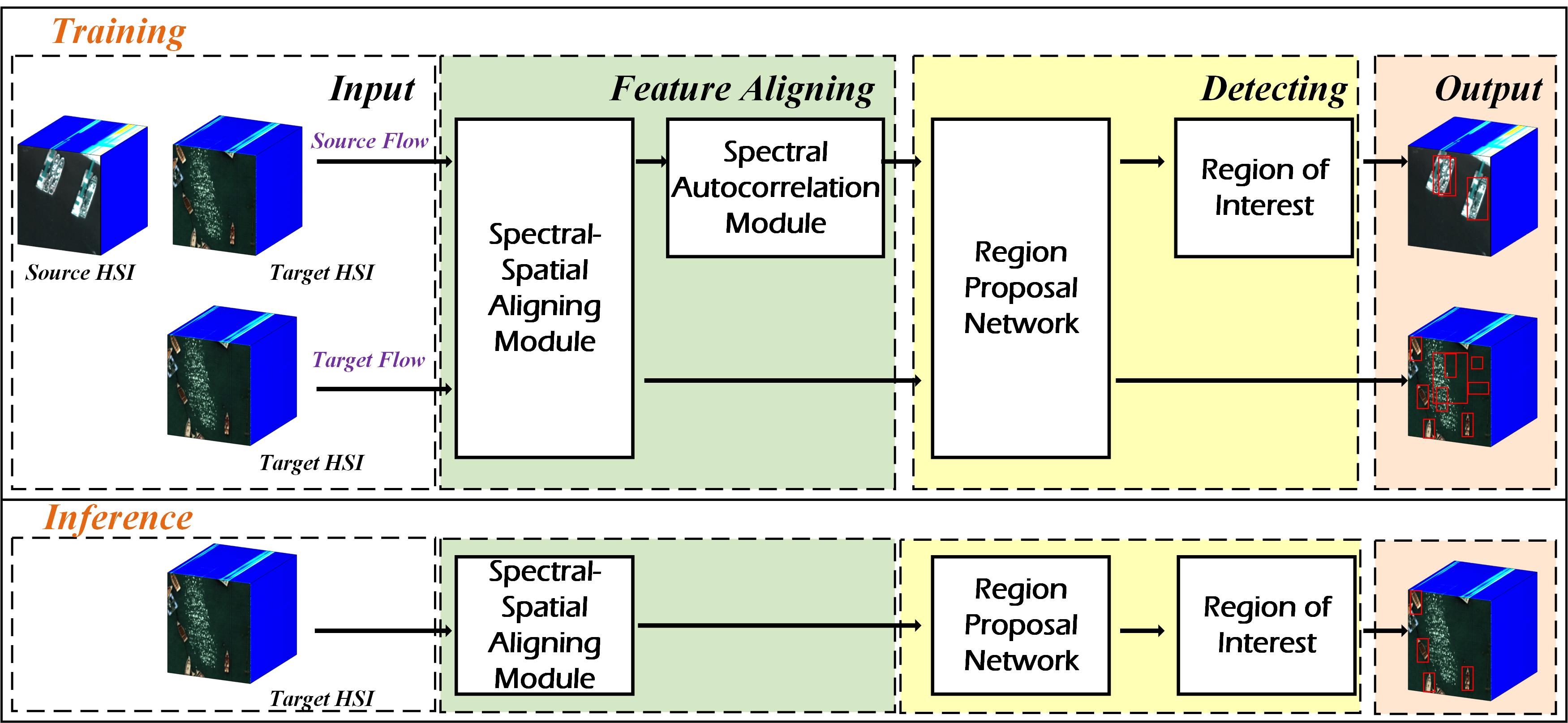}
\caption{The overview of SFA.}
\label{fig_SFA}
\end{figure*}

Although current HOD methods have achieved certain success, the effect of domain shift has not been investigated, which is the main focus of this manuscript.
\subsection{Unsupervised Cross-Domain Object Detection}
The UCOD task \cite{CSDA} can adapt the trained model from the labeled source to the unlabeled target domain. According to the manner of network branches, the UCOD can be roughly classified into single network-based and double network-based methods \cite{AT, HT, MRT, PT, D3T, CAT}, i.e., the teacher-student model. The single network-based methods have integrated adversarial feature learning to deceive the domain classifier. After that, the invariable features across domains have been extracted \cite{MAD, MGADA, Dafaster}. The double network-based methods attempt to generate pseudo-labels of target domain by a pre-trained teacher model from the source domain. Afterward, the pseudo-labels are fed into the student model for detection in the target domain \cite{AT, HT, MRT, PT}. Although existing methods have obtained perfect performance for UCOD tasks with RGB images, they mainly focus on the feature alignment in the spatial dimension, which cannot work on the HCOD task directly to tackle the domain shift in both the spectral and spatial domains.
\subsection{HSI Object Detection Datasets}
The HOD task has been developed for decades, but there are no sufficient large-scale public datasets now. The most popular one is the Airport-Beach-Urban (ABU) dataset \cite{ABU}, which only includes 13 HSIs of a rather small spatial size of 100×100. Although they have been widely used for HOD, its detected object is too obvious and its size is too small to train a deep model. Another popular one is the San Diego dataset, which suffers from the same problem of ABU. Hence, it is not ideal to develop deep learning-based HOD methods based on these above datasets. In 2023, Jang \textit{et al.} \cite{NIPS} proposed a large-scale dataset for detecting ships and sea plastic. However, there is still lacking of enough public HSI datasets to perform the HCOD task.
\section{Proposed Method}

\textbf{Notation} 
The HCOD datasets include a labeled source HSI dataset $\mathcal{D}_S = \left\{ (\boldsymbol{x}_i^S, \boldsymbol{y}_i^S)\right\}_{i=1}^{N_S}$, and an unlabeled target HSI dataset $\mathcal{D}_T = \left\{ \boldsymbol{x}_i^T \right\}_{i=1}^{N_T}$, in which the $\boldsymbol{x}_i\in{\mathbb{R}^{W \times H \times L}}$ represents the $i$-th HSI of width $W$, height $H$, and the number of bands $L$ in a dataset. The symbol $\boldsymbol{y}_i = \left\{ (\boldsymbol{b}_{i,j}, c_{i,j})\right\}_{j=1}^{N^{i}_{bbox}}$ is incorporated by the $j$-th bounding box coordinates $\boldsymbol{b}_{i,j}\in{x, y, w, h}$ and the class index $c_{i,j} \in {1,...,N_c}$. The $N_S$ and the $N_T$ stand for the number of the source HSI dataset and the target HSI dataset respectively.
\subsection{Overall Framework} 
Our proposed SFA method can be depicted in Figure \ref{fig_SFA}. Similar to \cite{Dafaster}, we have designed a single network-based framework with source and target flows. The training stage sequentially input source and target HSI into the source flow and the target HSI into the target flow, and both flows can be optimized simultaneously. As the source domain can provide the annotated label as the supervised signal but the target domain can only utilize the self-supervised signal, the training process of SFA may focus more on the source domain. To alleviate this problem, we utilize the target HSI twice in both flows to balance the feature learning of proposed framework.

The source flow process consecutively through the spectral-spatial alignment module (SSAM) and the spectral autocorrelation module (SACM), and an aligned local spectral-spatial feature can be obtained. The aligned feature is taken as an input to the region proposal network (RPN) \cite{RPN}, and the Region of Interest (ROI) module \cite{RPN} is utilized to achieve the detection result of the source HSI. On the contrary, the target flow only conducts the SSAM and the RPN as there are no bounding box label as the supervised signal.

\subsection{Loss Structure}\label{sec:loss}
For a better explanation of the SFA model, we firstly adopt a top-to-bottom way to clarify the loss structure of SFA. Specifically, the supervised loss consists of two parts, the loss from source and target flows are denoted by $L_S$ $L_T$, respectively. The overall optimization objective is given as:

\begin{equation}
L={{L}_{S}}+{{L}_{T}}
  \label{L}
\end{equation}

Source flow has simultaneously received the supervised signal by labeled HSI and the self-supervised signal from the target HSI. Accordingly, the $L_S$ is shown as:
\begin{equation}
{{L}_{S}}=\varepsilon {{L}_{s}^{r}}+\eta {{L}_{s}^{d}}+\tau {{L}_{SACM}}+{{L}_{s}^{rpn}}+{{L}_{roi}},
  \label{L_S}
\end{equation}
where the ${L}_{s}^{r}$ represents the reconstructed loss from the source flow and the symbols ${L}_{s}^{d}$ stands for the loss of the domain classifier from the source flow.

The ${L}_{SACM}$ indicates the loss of SACM. The ${{L}_{s}^{rpn}}$ and the ${{L}_{roi}}$ express the RPN loss from the source flow and the ROI loss respectively. It should be noted that the loss involved ${L}_{s}^{r}$, ${L}_{s}^{d}$, and ${{L}_{roi}}$ solely adopt the source HSI yet target HSI in the source flow. Among them, $\varepsilon$, $\eta$, and $\tau$ represent different default parameters to control the balance between different losses. In this manuscript, we have fixed them to be 0.5, 0.5, and 0.2, respectively.

Correspondingly, the target flow should only hold the self-supervised signal from the reconstruction error of target HSI. Therefore, the related loss can be expressed by the following formula:

\begin{equation}
{{L}_{T}}=\varepsilon {{L}_{t}^{r}}+\eta {{L}_{t}^{d}}+{{L}_{t}^{rpn}},
  \label{L_S}
\end{equation}
where the ${{L}_{t}^{r}}$ indicates the reconstructed loss from the target flow and the symbols ${L}_{t}^{d}$ stands for the loss of the domain classifier from the target flow. The ${{L}_{t}^{rpn}}$ expresses the RPN loss from the target flow.

The aforementioned loss lies in different modules from SFA. To sum up, the following part will provide a detailed description of the two designed modules. It should be noted that the number of bands across different domains is usually not the same. Therefore, it is essential to standardize the band dimensions to ensure consistency, allowing HSIs from source and target domain to input the SSAM together. To this end, we copy the first band and the last band of the source domain HSI and expand to the number of bands in the target domain HSI, when the HSI belonging to the source domain has fewer bands and the target domain image has more bands. If the target domain HSI contains fewer spectral bands, we will down-sample the spectral dimension of the source domain HSI to align with the target domain's band count. We assume that the target dataset or the source dataset after the band number matching process is still represented by $\mathcal{D}_T$ or the $\mathcal{D}_S$.

\subsection{Spectral-Spatial Aligning Module}
The SSAM aims to align the invariant features between domains. An underlying assumption is that the invariant features are more likely contained in the local spectral relationship between objects and their surrounding background and in the local spatial features like the shape and the texture information of the detected objects. Therefore, we should extract the local spectral-spatial features as the baseline for features aligning in the HCOD task. The AE model as a self-reconstructing mean can effectively investigate the hidden structure of HSI, where the hidden layer contains the key local spectral-spatial features to reform the original HSI. Accordingly, we propose the SSAM integrated with the AE to extract the invariant features between domains, and the domain classifier has been utilized to align the local spectral-spatial features.

\begin{figure}[htbp]
\centering
\includegraphics[width=0.5\textwidth, height=0.15\textheight]{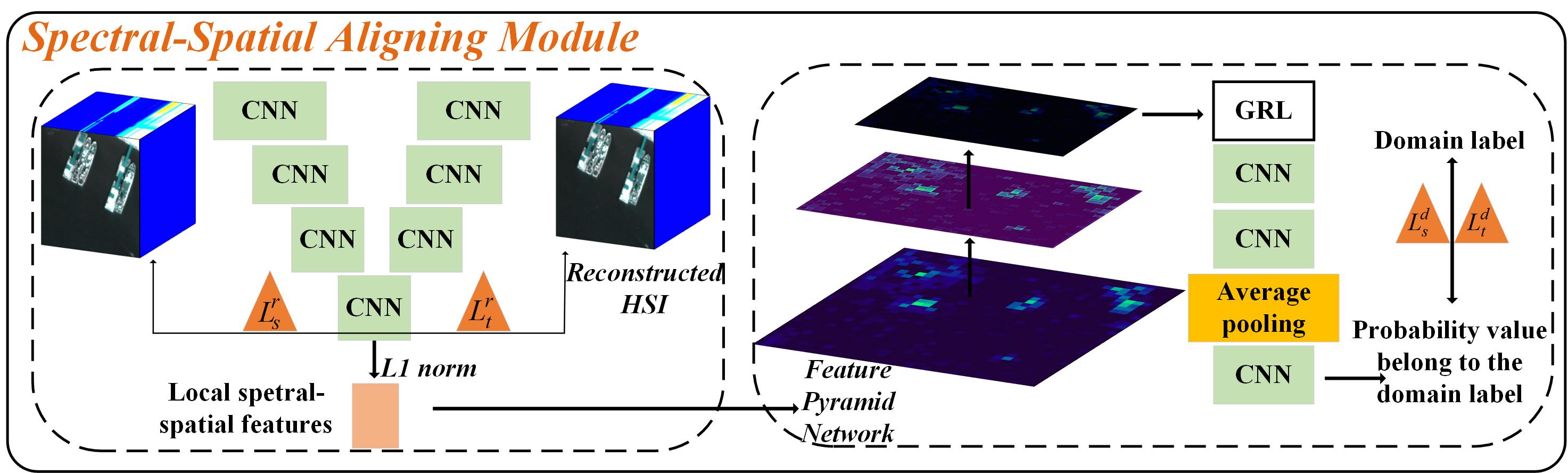}
\caption{The overview of the SSAM.}
\label{fig_SSAM}
\end{figure}

As shown in the Figure \ref{fig_SSAM}. Firstly, the HSI $\boldsymbol{x}_i$ can be fed to the encoder part of SSAM module, where the $L_1$ norm will be performed in the last layer of encoder. The reason is mainly caused by the local spectral-spatial features being less compared with features of inter-domain changes. We argue the $n$-level layer of the encoder as the $en_n$, and the operation of the AE is denoted by $ae(.)$. Accordingly, the reconstructed loss $L_{s}^{r}$ and the $L_{t}^{r}$ can be written as follows:
\begin{equation}
L_{s}^{r}=\left\| ae(\boldsymbol{x}_{i}^{S})-{\boldsymbol{x}_{i}^{S}} \right\|_{F}^{2}-+\alpha {{\left\| e{{n}_{3}}({\boldsymbol{x}_{i}^{S}}) \right\|}_{1}}
  \label{Lsr}
\end{equation}

\begin{equation}
L_{t}^{r}=\left\| ae(\boldsymbol{x}_{i}^{T})-{\boldsymbol{x}_{i}^{T}} \right\|_{F}^{2}-+\alpha {{\left\| e{{n}_{3}}({\boldsymbol{x}_{i}^{T}}) \right\|}_{1}}
  \label{Ltr}
\end{equation}

To obtain more multi-scale local spectral-spatial features beneficial for the HOD task, the obtained local spectral-spatial features from the above step are fed to the FPN. We denote the $FPN(en_3(\boldsymbol{x}_i))$ as the output from the FPN when inputting the $\boldsymbol{x}_i$. Afterwards, the domain classifier with sigmoid focus loss is performed to the third output layer of FPN to further unify the local spectral-spatial features. By linking a GRL before performing three $1\times1$ 2D convolutions and a $1\times1$ Average Pool (AVP), the output of domain classifier $DC(.)$ can be represented as follows.

\begin{equation}
{{D}_{s}^{o}}=DC(FP{{N}_{3}}(e{{n}_{3}}(\boldsymbol{x}_{i}^{S}))),
  \label{dso}
\end{equation}

\begin{equation}
{{D}_{t}^{o}}=DC(FP{{N}_{3}}(e{{n}_{3}}(\boldsymbol{x}_{i}^{T}))),
  \label{dto}
\end{equation}
where the ${D}_{s}^{o}$ stands for the output of the classifier from the source domain, i.e., the probability value belongs to the source. Accordingly, the ${D}_{t}^{o}$ stands for the output of the classifier from the target domain, i.e., the probability value belongs to the target. Therefore, the domain classifier loss can be equated as follows:

\begin{equation}
L_{s}^{d}=-\frac{(1-\lambda)}{\beta }\log [sigmoid(-\beta {{D}_{s}^{o}})],
  \label{Lsd}
\end{equation}

\begin{equation}
L_{t}^{d}=-\frac{\lambda}{\beta }\log [sigmoid(-\beta {{D}_{t}^{o}})],
  \label{Ltd}
\end{equation}

\begin{figure}[htbp]
\centering
\includegraphics[width=0.65\linewidth]{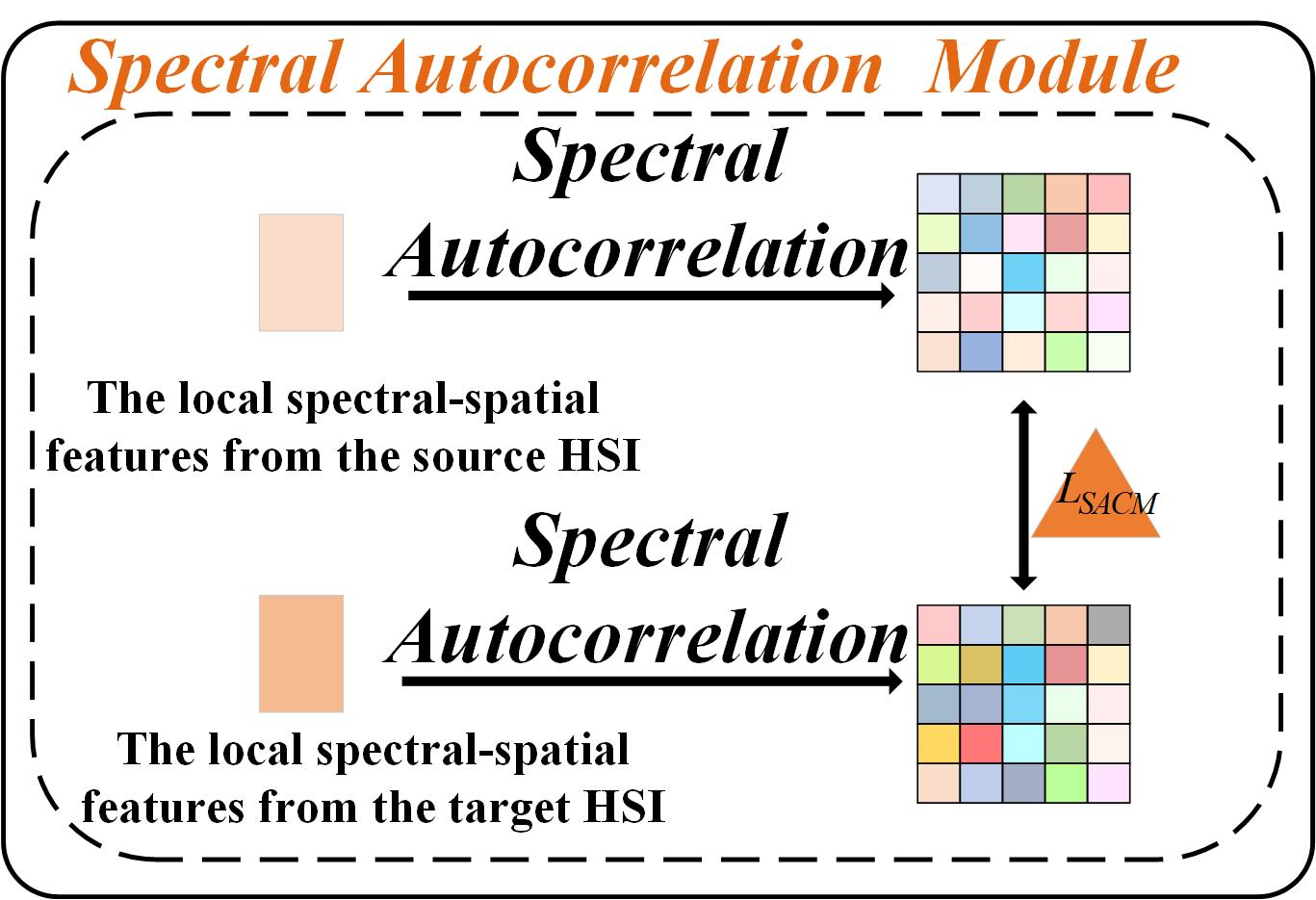}
\caption{The overview of the SACM.}
\label{fig_SACM}
\end{figure}

\begin{figure*}[htbp]
\centering
\begin{subfigure}{0.26\linewidth}
    \centering
    \includegraphics[width=\linewidth, height=4cm]{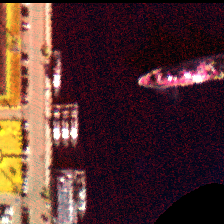}
    \caption{The M2SODAI dataset}
\end{subfigure}
\begin{subfigure}{0.26\linewidth}
    \centering
    \includegraphics[width=\linewidth, height=4cm]{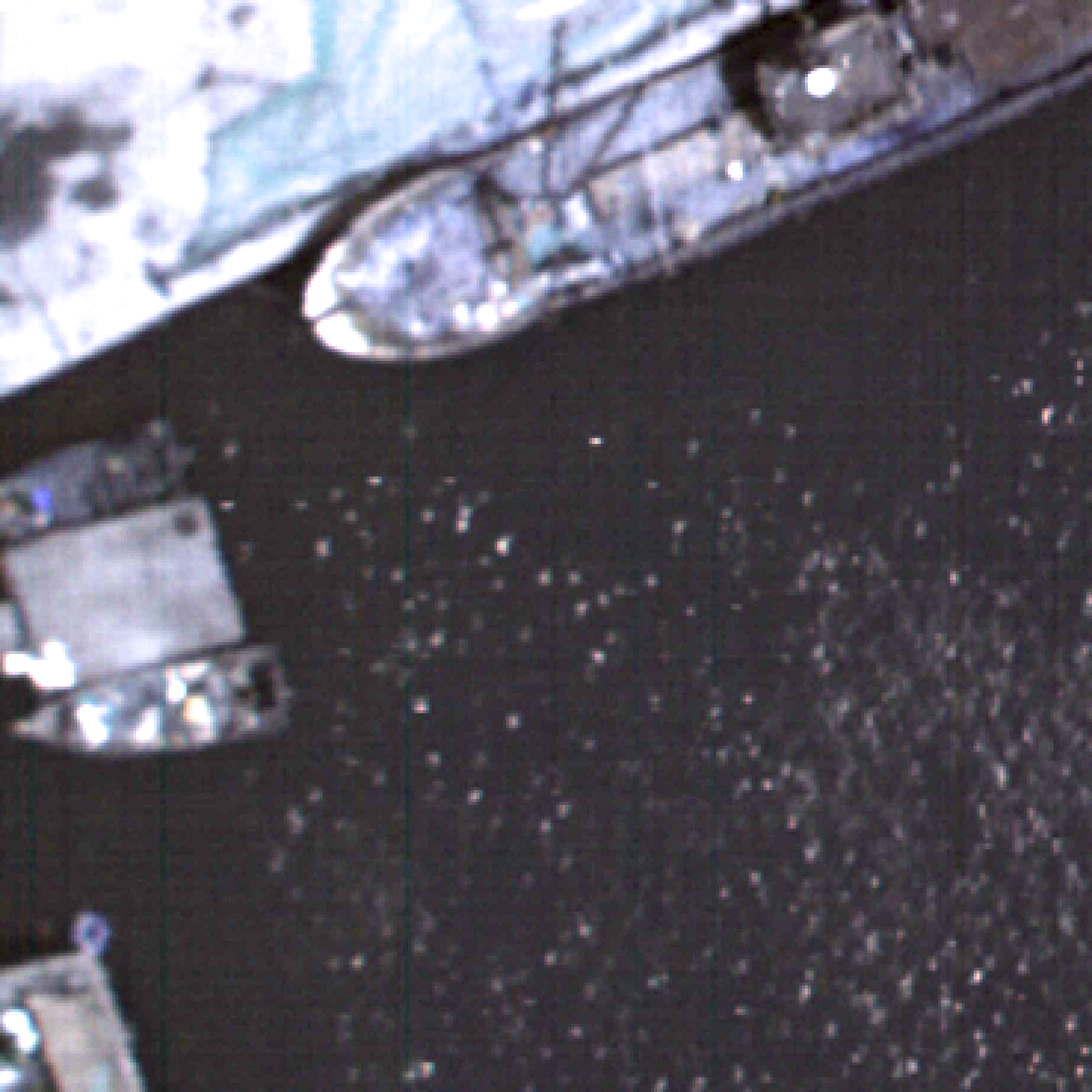}
    \caption{The LWP dataset}
\end{subfigure}
\begin{subfigure}{0.33\linewidth}
    \centering
    \includegraphics[width=\linewidth, height=4cm]{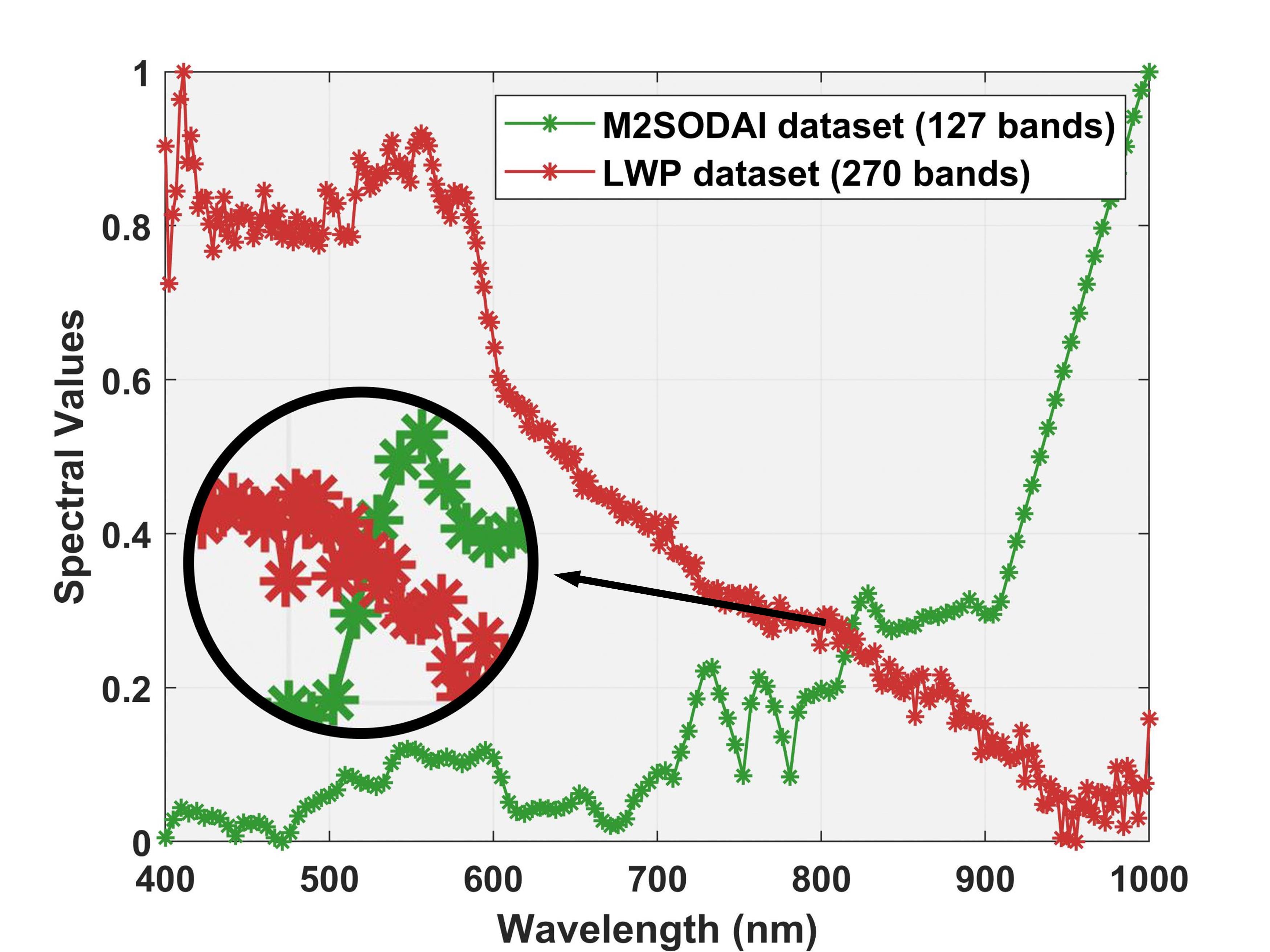}
    \caption{The mean spectral difference between datasets}
\end{subfigure}
\caption{Illustration of the domain shift phenomenon in spectral-spatial dimension between M2SODAI and LWP datasets. The M2SODAI dataset features a spatial resolution of 0.7$m$, as illustrated in (a), while the LWP dataset achieves a better spatial resolution up to 0.1$m$, as shown in (b). Figure (c) illustrates different spectral resolutions and shows that the sampling points in the spectral curve of LWP are denser, while the spectral curve of M2SODAI has a sparser distribution. This difference is attributed to the M2SODAI dataset's spectral resolution of 4.5$nm$, in contrast to a finer 2.22$nm$ resolution of the LWP dataset.
}
\label{fig_dataset}
\end{figure*}

Note that the $\beta$ and $\lambda$ represent default parameters and have been fixed to 2.0 and 0.25 separately. It should be noted that the classifier adopts GRL operation on the output of FPN. When propagating forward, its features remain unchanged. When performing the gradient back-propagation, the feature layer output by FPN is multiplied by a negative number (setting to -0.5 in the experiment) to obfuscate the local spectral-spatial features between domains. In this way, the minor differences from same kind of object will not be the problem.

The above losses involved $L_{s}^{r}$, $L_{s}^{r}$, $L_{s}^{d}$, and the $L_{t}^{d}$ constraints to the object detection model, and invariant local spectral-spatial characterization between domains can be excavated and aligned with the supervision. 

\subsection{Spectral Autocorrelation Module}
Compared to other UCOD task, the difficulty of HCOD is how to effectively align features from different domains with different spectral resolution. Therefore, the SACM shown in Figure \ref{fig_SACM} aims to further align the spectral features between domains. This module is based on an intrinsic hypothesis in the real situation that if the same kind of object are made of same material, the spectral features should be invariant between domains. 

In SACM, the source domain local spectral-spatial features $F_S=en_{3}(\boldsymbol{x}_i^S)$ and the target domain local spectral-spatial features $F_T=en_{3}(\boldsymbol{x}_i^T)$ are performing a spectral autocorrelation respectively. We then make these two spectral autocorrelation matrices as similar as possible supervised by the MSE loss. Accordingly, the loss $L_{SACM}$ which belongs to SACM can be shown as follows:
\begin{equation}
L_{SACM}=\left\| {{({{F}_{T}})}^{\mathbf{T}}}{{F}_{T}}-{{({{F}_{S}})}^{\mathbf{T}}}{{F}_{S}} \right\|_{F}^{2}
  \label{Ls}
\end{equation}
\begin{algorithm}
	\caption{The proposed SFA.}
	\label{hcod_algorithm}
	\begin{algorithmic}[1] 	
		\STATE \textbf{Begin}
		\vspace{0.01cm}
            \STATE \textbf{Training}
		\vspace{0.01cm}
		\WHILE{training process}
		\vspace{0.01cm}
		\STATE $(\boldsymbol{x}_i^S, \boldsymbol{y}_i^S)$ and $\boldsymbol{x}_i^T$ are fed to the SFA;;
		\vspace{0.01cm}
		\STATE Extracting the local spectral-spatial features from the source flow by Eq.(\ref{Lsr});
		\vspace{0.01cm}
		\STATE Aligning the spectral features by Eq.(\ref{Ls});
		\vspace{0.01cm}
		\STATE Inputting the local spectral-spatial features to the FPN and aligning them by Eq.(\ref{dso}) and Eq.(\ref{Lsd});
		\vspace{0.01cm}
		\STATE Inputting the output features of FPN to the RPN and the ROI;
		\vspace{0.01cm}
		\STATE $\boldsymbol{x}_i^T$ is fed to the SFA;
		\vspace{0.01cm}
		\STATE Extracting the local spectral-spatial features by Eq.(\ref{Lsd});
		\vspace{0.01cm}
            \STATE Inputting the output features of FPN to the RPN;
            \vspace{0.01cm}
            \STATE The SFA is optimized by Eq.(\ref{L}).
		\vspace{0.01cm}
		\ENDWHILE
            \vspace{0.01cm}
            \STATE \textbf{Inference}
		\vspace{0.01cm}
            \STATE $\boldsymbol{x}_i^T$ is fed to the trained SFA;
		\vspace{0.01cm}
		\STATE $\boldsymbol{x}_i^T$ is fed to the trained SSAM;.
		\vspace{0.01cm}
            \STATE the acquired data, i.e., the output of FPN,  by the previous step is fed to the trained RPN;.
		\vspace{0.01cm}
            \STATE the proposals from the previous step are conducted with trained ROI;
		\vspace{0.01cm}
            \STATE Generate the detection results of target HSIs.
		\vspace{0.01cm}
		\STATE \textbf{End}
	\end{algorithmic}
\end{algorithm}
 
 Next, we need to input the acquired features $FPN(en_3(\boldsymbol{x}_i))$ instead of the $FPN_3(en_3(\boldsymbol{x}_i))$ into the RPN and the ROI in succession due to it belong to the source flow. To sum up, the overall step of SFA can be summarized in Algorithm~\ref{hcod_algorithm}.

\begin{table*}[htbp]
  \centering
  \caption{Quantitative Experimental Results for Comparative Experiments.}
    \resizebox{0.95\textwidth}{!}{
    \begin{tabular}{ccccccccc}
    \toprule
    &AP&AP$\_small$&AP$\_midium$ &AP$\_large$&AR&AR$\_small$&AR$\_meduim$&AR$\_large$ \\
    \midrule
    Upper boundary & 42.18\% & 12.86\% & 18.75\% & 13.82\% &25.71\%& 17.00\% & 29.56\% & 24.16\% \\
    DA-Faster (CVPR,2018) \cite{Dafaster} & 0.02\% & 0.00\% & 0.00\% & 0.63\%&2.35\% & 1.44\% & 2.45\% & 7.50\% \\
    PT (IMCL, 2022) \cite{PT} & 0.04\% & 0.00\% & 0.00\% & 0.66\% &1.51\%& 0.11\% & 0.57\% & \textbf{28.33\%} \\
    MGADA (CVPR, 2022) \cite{MGADA} & 0.95\% & 0.06\% & 2.03\% & \textbf{2.57\%} & 9.00\%&6.55\% & 10.48\% & 2.50\% \\
    AT (CVPR, 2022) \cite{AT} & 0.02\% & 0.00\% & 0.01\% & 0.00\% &2.06\%& 2.88\% & 1.58\% & 4.16\% \\
    MRT (ICCV, 2023) \cite{MRT} & 0.00\% & 0.00\% & 0.00\% & 0.00\% & 0.00\%& 0.00\% & 0.00\% & 0.00\% \\
    DoubleFPN (NIPS, 2023) \cite{NIPS} & 0.06\% & 0.04\% & 0.10\% & 0.00\% & 0.12\%& 0.44\%  & 0.00\% & 0.00\% \\
    Ours  & \textbf{23.93\%} & \textbf{4.38\%} & \textbf{16.87\%} & 0.00\% & \textbf{30.00\%} & \textbf{33.88\%} & \textbf{30.04\%} & 0.00\% \\
    \bottomrule
    \end{tabular}
    }
  \label{QE}%
\end{table*}%

\subsection{Longwang Port Datasets}
To achieve a fair experimental comparison, constructing the cross-domain datasets of HSI is essential. We have collected and annotated an HSI dataset named Longwang Port (LWP). It is acquired in Dalian, Liaoning Province, China, which mainly consist of fishing ships made of different materials and kelp ships made of wood. We use an HSI sensor to capture three different spatial resolutions HSI of 0.1$m$, 0.15$m$, and 0.2$m$. The HSI sensor is the Nano-Hyperspec designed by Headwall Photonics, which is a new miniature airborne hyperspectral imager specifically developed for UAV airborne platforms. It has a spectral range of 400-1000$nm$, a spectral channel count of 270, and a spectral sampling rate of about 2.2$nm$. Examples of collected LWP dataset therein can be visualized in Figure \ref{fig_dataset}. 

To enable the datasets for the HCOD task, we have annotated the LWP dataset and split them into a large-scale datasets in total of 1634 HSIs, and the spatial size 224$\times$224 (same with M2SODAI dataset), in which the training datasets consisted of 1339 HSIs with 1790 ships, the validation datasets consisted of 164 HSIs with 310 ships, and the test datasets consisted of 134 HSIs with 119 ships. As the M2SODAI contains 1257 HSIs with a 4.5$nm$ spectral resolution and a 0.7$m$ spatial resolution, we have collected a much larger HSI dataset with 1634 HSIs a much better spectral resolution about 2.2$nm$ and spatial resolution about 0.1-0.2$m$. As a result, the M2SODAI datasets \cite{NIPS} are combined with LWP leading to the HCOD dataset. The corresponding schematic diagram is shown in Figure \ref{fig_dataset}.

\section{Experiments}

\subsection{Implementation and Evaluation Metrics } \label{sec:implementation_evaluation}
\noindent\textbf{Dataset Arrangement.} 
The M2SODAI dataset is used as the source domain dataset, where its object class contains the ship and the plastics in the original one. However, we only focus on the ship object to match with the LWP dataset which only ships existed. The LWP dataset is adopted as the target domain. It should be noted that the source dataset has to be in the same spectral range as the target dataset because of retains the possibility of spectral matching. For the LWP dataset, the validation dataset with annotations is used to assess the HCOD performance. For the training set of LWP in HCOD, the annotations are ignored. In this manuscript, we have defined the small, medium, and large objects with a spatial size of $<$32$^2$, 32$^2$-95$^2$, and $\geq$96$^2$, respectively.

\noindent\textbf{Comparison methods.} As this is the first proposed method for the HCOD task, there are no specific compared methods yet. As a result, we compare the existing state-of-the-art (SOTA) UCOD methods concerning normal RGB images. Specifically, we compare with the single network-based method, such as Dafaster \cite{Dafaster}, MGADA \cite{MGADA}, the double network-based method such as PT \cite{PT}, AT\cite{AT}, and MRT \cite{MRT}. The DoubleFPN \cite{NIPS} as the first work of leveraging the multiple spectral priors is not a HCOD-based method, but we compare with it to further validate the essential of HCOD. To know the upper boundary of HCOD, we also utilize the DoubleFPN to train and test in the target domain.

\noindent\textbf{Implementation details.} The SFA model is implemented based on the detectron2 \cite{wu2019detectron2} framework, and the Average Precision (AP) with an IoU threshold of 0.5 is adopted as the quantitative metric. To report the detection performance of objects with different sizes, we have adopted AP with an IoU threshold ranging from 0.5 to 0.95. The setting of the comparison methods remains unchanged except for the number of input channels of the backbone. To enable the comparison methods use HSIs from both domains, we performed band expansion on the source domain HSI, as shown in the last paragraph of the Sec \ref{sec:loss}. We set the batch-size to be 2 for the source and target domain with an RTX 1080ti GPU. The SFA model is optimized with an Adam optimizer of a 3e-4 learning rate without any decay, and the number of training iteration is set to be 15000. 
\begin{figure*}[htbp]
    \centering
    \begin{subfigure}{0.24\textwidth}
        \centering
        \includegraphics[width=\textwidth]{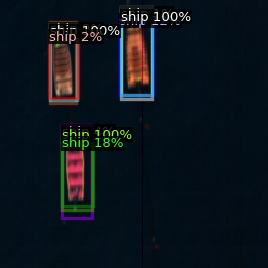}
        \caption{Upper boundary}
    \end{subfigure}
    \hfill
    \begin{subfigure}{0.24\textwidth}
        \centering
        \includegraphics[width=\textwidth]{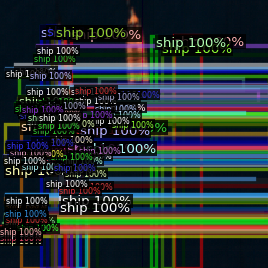}
        \caption{Da-Faster}
    \end{subfigure}
    \hfill
    \begin{subfigure}{0.24\textwidth}
        \centering
        \includegraphics[width=\textwidth]{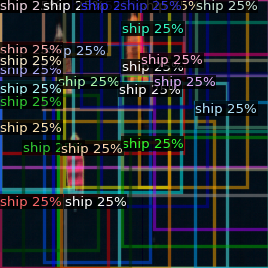}
        \caption{PT}
    \end{subfigure}
    \hfill
    \begin{subfigure}{0.24\textwidth}
        \centering
        \includegraphics[width=\textwidth]{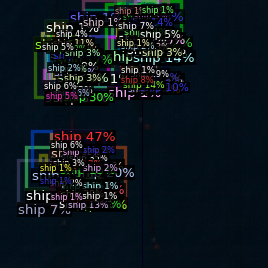}
        \caption{MGADA}
    \end{subfigure}
    
    \vspace{0.5cm}
    
    \begin{subfigure}{0.24\textwidth}
        \centering
        \includegraphics[width=\textwidth]{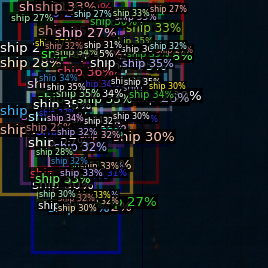}
        \caption{AT}
    \end{subfigure}
    \hfill
    \begin{subfigure}{0.24\textwidth}
        \centering
        \includegraphics[width=\textwidth]{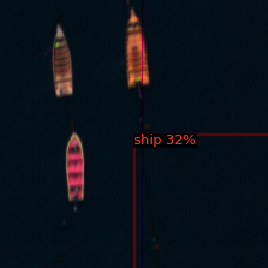}
        \caption{MRT}
    \end{subfigure}
    \hfill
    \begin{subfigure}{0.24\textwidth}
        \centering
        \includegraphics[width=\textwidth]{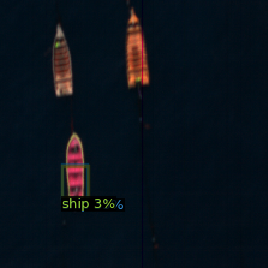}
        \caption{DoubleFPN}
    \end{subfigure}
    \hfill
    \begin{subfigure}{0.24\textwidth}
        \centering
        \includegraphics[width=\textwidth]{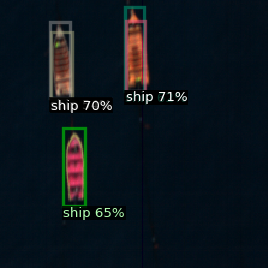}
        \caption{Ours}
    \end{subfigure}
    
    \caption{Qualitative results on the target domain.}
    \label{fQ}
\end{figure*}

\subsection{Comparison results.} \label{sec:CR}
The performance of the SFA approach and the recent SOTA UCOD methods are reported in Table \ref{QE}. We find that our proposed method obtains the best AP of 23.9\%. By contrast, the others seems to be failed with AP around 0. For the DA-Faster \cite{Dafaster} with a VGG-16 backbone, the spectral features are damaged due to the lacking of introducing more band feature. The DA-Faster detector cannot effectively extract spatial-spectral features of the source and target domains. The PT model \cite{PT} is based on the pseudo-label generation, which over-rely on the quality of the pseudo-label from the target domain. As mentioned above, the spectral-spatial features failed to be extracted. During the experiment, pseudo labels cannot be used at all because their scores are too low like only 0.25 in any instances as shown in the Figure. \ref{fQ}. Thus, the accordingly pseudo-label can not be generated in the target domain and the HCOD by PT is not succeed.

Regarding MGADA \cite{MGADA}, it adopts the pixel, instance and category-level information to align the features in both domains. With multiple level information, the ability to extract the features improves a bit to locate objects. However, the HSI holds more detailed spectral features than the RGB image, the MGADA method has not suitable for HCOD and the detect results are rather chaotic. For AT \cite{AT}, the detection performance is very similar to MGADA. With regard to the MRT \cite{MRT}, the predicted bounding box is not generated due to the teacher model itself failing to adapt to the spectral-spatial features. The DoubleFPN \cite{NIPS} is specially designed for the HSI with a promising feature extraction capability in the HOD task. However, the spectral-spatial features are shifted between domains in the HCOD task, the performance of DoubleFPN approaches is not as ideal as its performance in the HOD. 

For our SFA method, it still has a significant gap in quantitative indicators compared to the upper boundary. As shown in the visualization map in Figure \ref{fQ}, the upper boundary uses more bounding boxes to locate the object, which is the reason for its higher accuracy. Although numerous bounding boxes contain the target objects in the detection results of compared methods, they are either too large or too small, resulting in an IoU that does not exceed 0.5, Hence, all the compared methods can be assumed as failure and their AP are around 0. 

To better indicate the performance of these SOTA methods, we have introduced the average recall (AR) with an IOU threshold ranging from 0.5 to 0.95 in our experiments, which is shown in Table \ref{QE}. However, the ARs of these SOTA methods are still lower than our proposed method, which again indicates the detection performance of our proposed SFA. For SFA, the AR is higher than the upper boundary, which indicates the strong ability of locating detects of our method. Although the AR$\_large$ of SFA is lower than PT, the AP of PT is much lower, where it has more false alarm detection. In the full consideration of AP and AR, our method still achieves the best detection performance 
\subsection{Ablation study}
\begin{figure}[t]
\centering
\begin{subfigure}{0.4\linewidth}
    \centering
    \includegraphics[width=\linewidth]{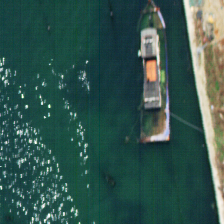}
    \caption{The original HSI}
\end{subfigure}
\hspace{0.05\linewidth} 
\begin{subfigure}{0.4\linewidth}
    \centering
    \includegraphics[width=\linewidth]{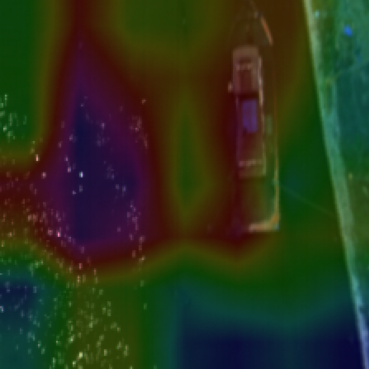}
    \caption{SFA w/o SSAM+SACM}
\end{subfigure}
\vspace{0.5em} 
\begin{subfigure}{0.4\linewidth}
    \centering
    \includegraphics[width=\linewidth]{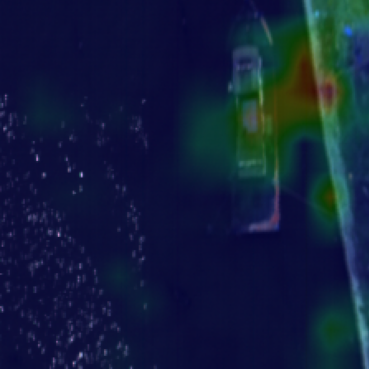}
    \caption{SFA w/o SACM}
\end{subfigure}
\hspace{0.05\linewidth} 
\begin{subfigure}{0.4\linewidth}
    \centering
    \includegraphics[width=\linewidth]{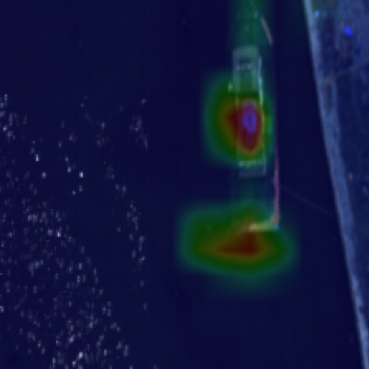}
    \caption{SFA}
\end{subfigure}

\caption{The feature map for the ablation experiment.}
\label{fig_AE}
\end{figure}

To validate the effectiveness of designed modules containing SSAM and SACM, we conduct an ablation study as recorded in Table \ref{tab_as}. We retrain the model under unchanged experimental conditions and parameters by replacing the SSAM with Resnet50 assembled the FPN module and removing the SACM, which is denoted by SFA w/o SSAM+SACM. We find a sharp decline in AP. As shown in Figure \ref{fig_AE}, the corresponding feature map (b) shows a diffuse phenomenon, and it can be seen that both the object and its surrounding background has been focused, which can be assumed as a failure for the HCOD. This indicates that the features extracted by SFA are effective for HCOD, while the rest are not suitable. 

To demonstrate the effectiveness of the SSAM, we remove the SACM and validate the performance. Compared with the Resnet50 with the FPN, the SSAM improves 15.8\% AP, indicating the SSAM plays a key role in the SFA model. The corresponding feature map in Figure \ref{fig_AE} (c) displays that the SSAM has the ability to reduce the false alarm in background and improve the detection performance. However, extracted features still focus part of the background, which is not fully desired in the HCOD task.
\begin{table}[htbp]
  \centering
  \caption{Quantitative Results for Ablation Studies.}
  \resizebox{0.22\textwidth}{!}{
    \begin{tabular}{cc}
    \toprule
          & AP \\
    \midrule
    SFA w/o SSAM+SACM & 1.1\% \\
    SFA w/o SACM & 16.9\% \\
    SFA   & \textbf{23.9\%} \\
    \bottomrule
    \end{tabular}%
    }
  \label{tab_as}%
\end{table}%

Afterwards, we add the SACM module to compose the SFA model, the accuracy further improves to 23.9\%. This suggests that the SACM module is effective for spectral alignment. The feature map shows the structure of ship are marked as red. In other words, the difference between the object and the neighboring background is more remarkable with the aid of SFA. Thus, the SFA is capable of capturing such invariant characteristics between different domains. As SACM is highly related with SSAM, the SACM cannot be performed without SSAM, we do not conduct the ablation study by comparing with SFA w/o SSAM.

\begin{figure}[t]
\centering
\includegraphics[width=0.65\linewidth]{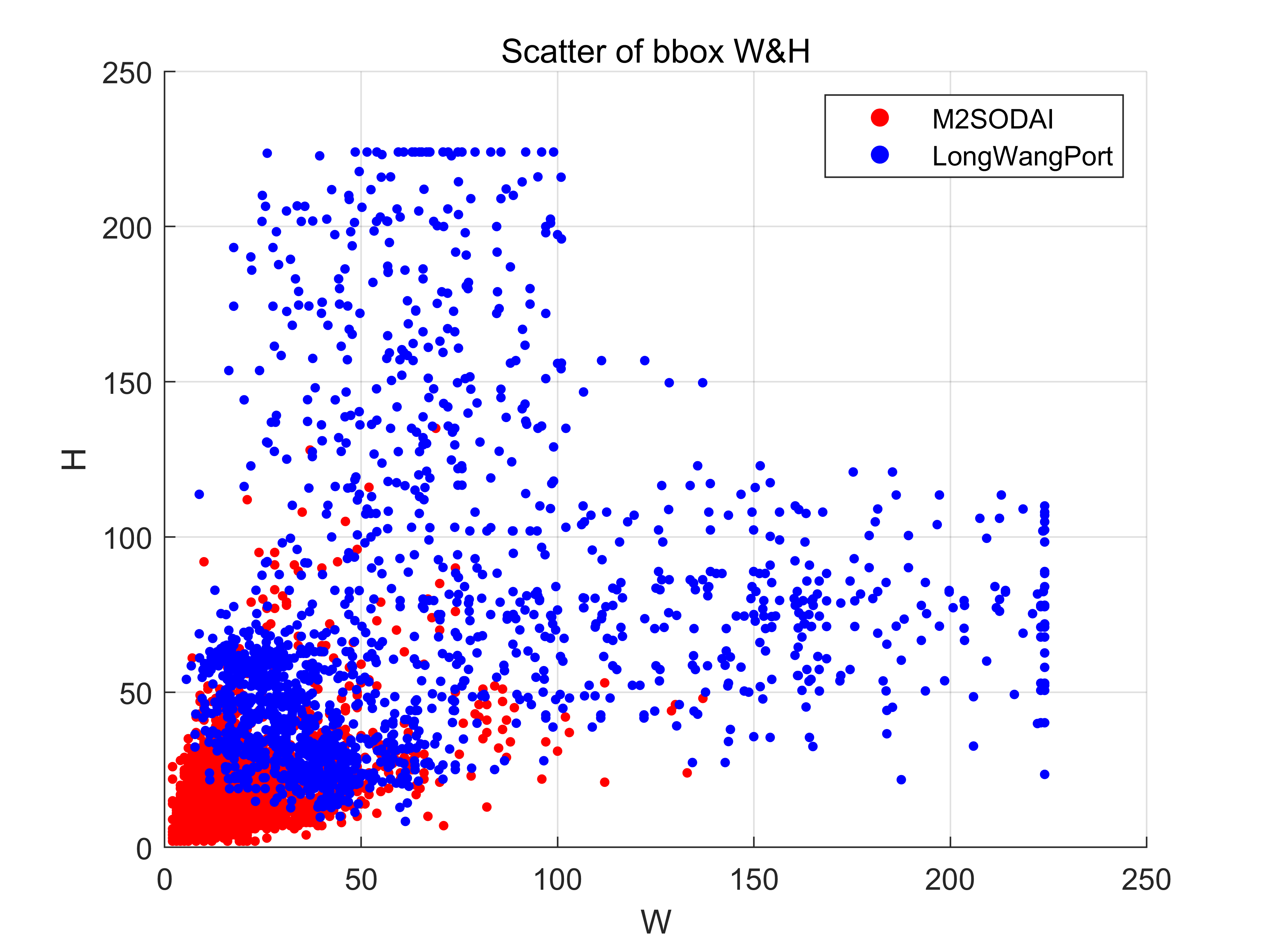} 
\caption{The bounding box distribution between M2SODAI \cite{NIPS} and LWP datasets.}
\label{fig_bbox}
\end{figure}

\subsection{Discussion} \label{sec:D}
Although we have achieved the HCOD by our proposed SFA, we find the main contribution of the AP is from both the small and medium objects. To find out the reason of this phenomenon, we have visualized the bounding box distribution maps of both the source and target domain datasets as shown in Figure \ref{fig_bbox}. We have observed that within the size range of 25$^2$-50$^2$, the source and target domains are very similar. In other words, the medium size is mainly located in this size range, and the rest are small objects. Accordingly, the AP\_medium archives the performance approaching the Upper boundary. Finally, how to decouple the object size from the spectral-spatial features is the key to further improve the performance of HCOD.

\section{Conclusion}
In this paper, we have proposed the SFA model for the HCOD task, which is the first one in object detection community. A robust HSI detector for a target domain without any additional labeled data can be acquired based on SFA. Our key observation is that the local spectral-spatial features are invariant between the source and the target domain. Accordingly, we have designed the SSAM to alleviate the domain shift within spectral-spatial dimension, and the SACM is developed to further aligning the spectral relationship. The experimental results have validated the advantage of SFA over current UCOD SOTA approaches. In the near future, we will release the LWP dataset and attempt to collect more useful datasets for the advancement of HCOD.

{
    \small
    \bibliographystyle{ieeenat_fullname}
    \bibliography{main}
}
\clearpage
\setcounter{page}{1}
\maketitlesupplementary

\setcounter{table}{0} 

\begin{table*}[htbp]
  \centering
  \caption{Quantitative Experimental Results for Comparative Experiments of LWP $\to$ M2SODAI.}
  \resizebox{0.95\textwidth}{!}{
    \begin{tabular}{ccccccccc}
    \toprule
          & AP & AP\_small & AP$\_midium$ & AP$\_large$ & AR   & AR$\_small $& AR$\_meduim $& AR$\_large$ \\
    \midrule
    Upper boundary & 22.72\% & 6.25\% & 31.25\% & N/A   & 14.79\% & 13.87\% & 45.33\% & N/A \\
    DA-Faster (CVPR,2018) \cite{Dafaster} & 0.00\% & 0.00\% & 0.00\% & N/A   & 0.05\% & 0.06\% & 0.00\% & N/A \\
    PT (IMCL, 2022) \cite{PT} & 0.00\% & 0.00\% & 0.00\% & N/A   & 0.00\% & 0.00\% & 0.00\% & N/A \\
    MGADA (CVPR, 2022) \cite{MGADA} & 0.35\% & 0.33\% & 4.11\% & N/A   & \textbf{2.17\%} & \textbf{1.23\%} & 33.30\% & N/A \\
    AT (CVPR, 2022) \cite{AT} & 0.14\% & 0.02\% & 0.80\% & N/A   & 0.78\% & 0.34\% & 15.33\% & N/A \\
    MRT (ICCV, 2023) \cite{MRT} & 0.00\% & 0.00\% & 0.00\% & N/A   & 0.00\% & 0.00\% & 0.00\% & N/A \\
    DoubleFPN (NIPS, 2023) \cite{NIPS} & 0.00\% & 0.00\% & 0.00\% & N/A   & 0.03\% & 0.04\% & 0.00\% & N/A \\
    Ours  & \textbf{1.71\%} & \textbf{0.37\%} & \textbf{12.24\%} & N/A   & 2.07\% & 0.70\% & \textbf{47.33\%} & N/A \\
    \bottomrule
    \end{tabular}%
    }
  \label{tab_s}%
\end{table*}%

\section*{A. Additional Experiment}
\label{sec:rationale}

To further validate the effectiveness of the proposed SFA,  we conduct another cross-domain setting, i.e., from the LWP dataset to the M2SODAI dataset. The corresponding quantitative results are shown in Tab. \ref{tab_s}. It is similar to the experimental setting in the manuscript that performing the cross-domain object detection by using M2SODAI as the source domain, the comparison method is significantly declined even invalid. Our method achieves the best performance compared to other SOTA methods. This is mainly attributed to the medium-sized objects, which achieve an AP of 12.24\%. As there is no large object in the M2SODAI dataset, the AP$\_large$ is not applicable in the experiment. Although the AP is not outstanding, the AR is promising even higher than the upper boundary in the AR$\_meduim$. This indicates that the SFA model is strongly capable of detecting medium objects where both exist on both datasets. In addition, the visualization of detection results is shown in the Fig. \ref{f_S}. It can be found the a huge spatial difference in contrast with LWP datasets. 

\setcounter{figure}{0} 
\begin{figure}[htbp]
    \centering
    \begin{subfigure}{0.2\textwidth}
        \centering
        \includegraphics[width=\linewidth]{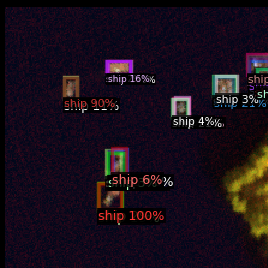}
        \caption{Upper boundary}
    \end{subfigure}
    \hfill
    \begin{subfigure}{0.2\textwidth}
        \centering
        \includegraphics[width=\linewidth]{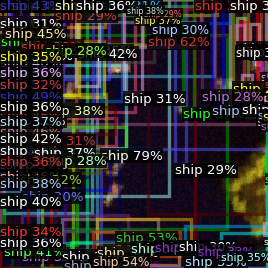}
        \caption{Da-Faster}
    \end{subfigure}
    
    \vspace{0.3cm}
    
    \begin{subfigure}{0.2\textwidth}
        \centering
        \includegraphics[width=\linewidth]{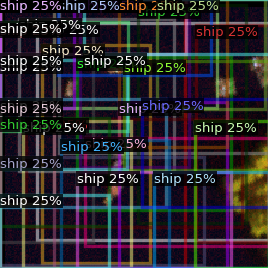}
        \caption{PT}
    \end{subfigure}
    \hfill
    \begin{subfigure}{0.2\textwidth}
        \centering
        \includegraphics[width=\linewidth]{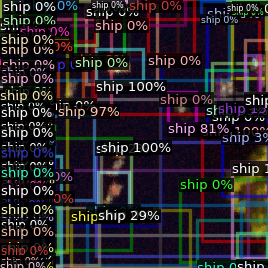}
        \caption{MGADA}
    \end{subfigure}
    
    \vspace{0.3cm}
    
    \begin{subfigure}{0.2\textwidth}
        \centering
        \includegraphics[width=\linewidth]{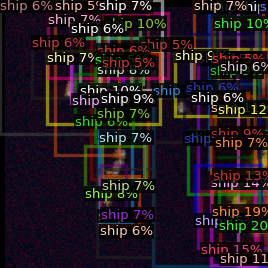}
        \caption{AT}
    \end{subfigure}
    \hfill
    \begin{subfigure}{0.2\textwidth}
        \centering
        \includegraphics[width=\linewidth]{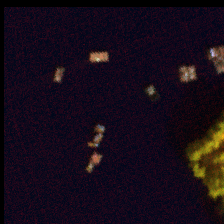}
        \caption{MRT}
    \end{subfigure}
    
    \vspace{0.3cm}
    
    \begin{subfigure}{0.2\textwidth}
        \centering
        \includegraphics[width=\linewidth]{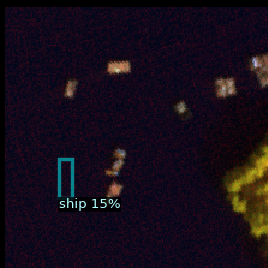}
        \caption{DoubleFPN}
    \end{subfigure}
    \hfill
    \begin{subfigure}{0.2\textwidth}
        \centering
        \includegraphics[width=\linewidth]{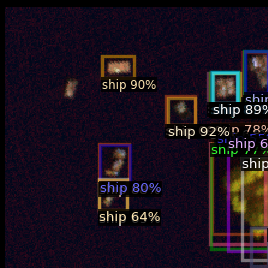}
        \caption{Ours}
    \end{subfigure}
    
    \caption{Qualitative results on the LWP $\to$ M2SODAI.}
    \label{f_S}
\end{figure}

For the performance of LWP $\to$ M2SODAI is not as good as the M2SODAI $\to$ LWP, it is mainly caused by the various differences in the spatial dimension. In M2SODAI, it holds a 0.7$m$ spatial resolution, and the scene is captured within a larger region compared 
to the LWP with a higher spatial resolution. Furthermore, the larger region implies that a more complicated background. As shown in Fig. \ref{f_S} (h), the land background on the right side is mistakenly identified as part of the ship object. When using the M2SODAI as the source domain, we can leverage the corresponding annotation. Therefore, it enables the method to adapt to the sophisticated scene, and it can also be backward compatible with simple scenarios. In contrast, the LWP acting as the source domain is simpler, the generalization from a simple source to a difficult target domain is more challenging. Accordingly, cross-domain object detection from a higher spatial resolution LWP to a lower spatial resolution M2SODAI cannot achieve satisfactory performance. In future work, we will attempt to solve this problem.

\end{document}